\documentclass[11pt,a4paper]{article}
\usepackage[hyperref]{acl2020}
\usepackage{times}
\usepackage{latexsym}

\usepackage{amsfonts}
\usepackage{url}
\usepackage{xcolor}
\usepackage[ruled,linesnumbered]{algorithm2e}
\usepackage{graphicx}
\usepackage{bbm}

\usepackage{amsmath}

\usepackage{microtype}

\aclfinalcopy 

\title{DomBERT: Domain-oriented Language Model \\for Aspect-based Sentiment Analysis}

\author{Hu Xu\textsuperscript{\text{1}}, Bing Liu\textsuperscript{\text{1}}, Lei Shu\textsuperscript{\text{1}}\and Philip S. Yu\textsuperscript{\text{1,2}}\\
    \textsuperscript{1}{Department of Computer Science, University of Illinois at Chicago, Chicago, IL, USA}\\
    \textsuperscript{2}{Institute for Data Science, Tsinghua University, Beijing, China}\\
    {\tt \{hxu48, liub, lshu3, psyu\}@uic.edu}
}

\date{}

\begin{document}
\maketitle
\begin{abstract}
This paper focuses on learning domain-oriented language models driven by end tasks, which aims to combine the worlds of both general-purpose language models (such as ELMo and BERT) and domain-specific language understanding.
We propose DomBERT, an extension of BERT to learn from both in-domain corpus and relevant domain corpora.
This helps in learning domain language models with low-resources.
Experiments are conducted on an assortment of tasks in aspect-based sentiment analysis, demonstrating promising results.
\footnote{Work in progress.}.
\end{abstract}

\section{Introduction}
Pre-trained language models (LMs) \cite{peters2018deep,radford2018improving,radford2019language,devlin2018bert} aim to learn general (or mixed-domain) knowledge for end tasks. Recent studies \cite{xu_bert2019,dontstoppretraining2020} show that learning domain-specific LMs are equally important because general-purpose LMs lack enough focus on domain details.
This is partially because the training corpus of general LMs is out-of-domain for domain end tasks and, more importantly, 
because mixed-domain weights may not capture the long-tailed and under represented domain details \cite{xu_acl2018} (see Section \ref{sec:exp}).
An intuitive example can be found in Table \ref{tbl:example}, where all masked words \textit{sky}, \textit{water}, \textit{idea}, \textit{screen} and \textit{picture} can appear in a mixed-domain corpus. A general-purpose LM may favor frequent examples and ignore long-tailed choices in certain domains.

\begin{table}
    \centering
    \scalebox{0.8}{
        \begin{tabular}{l|l}
            \hline
            Example & Domain \\
            \hline\hline
            The $\texttt{[MASK]}$ is clear . \\
            \hline\hline
            The \textit{sky} is clear . & Astronomy [Irrelevant Domain]\\
            The \textit{water} is clear . & Liquids [Irrelevant Domain]\\
            The \textit{idea} is clear . & Concepts [Irrelevant Domain]\\
            The \textit{screen} is clear . & Desktop [Relevant Domain]\\
            \hline
            The \textit{picture} is clear . & Laptop [Target Domain]\\
            \hline
        \end{tabular}
    }
    \caption{Multiple choices of a masked token from different domains lead to confusing ground-truths and the hardness of finding a general updates of parameters in masked language model.}
    \label{tbl:example}
    \vspace{-5mm}
\end{table}

In contrast, although domain-specific LMs can capture fine-grained domain details, they may suffer from insufficient training corpus \cite{dontstoppretraining2020} to strengthen general knowledge within a domain.
To this end, we propose a domain-oriented learning task that aims to combine the benefits of both general and domain-specific world:

\noindent \textbf{Domain-oriented Learning}: Given a target domain $t$ and a set of diverse source domains $S=\{s_1, s_2, \dots\}$, perform (language model) learning that focusing on $t$ and all its relevant domains in $S$.

This learning task resolves the issues that exist in both general and domain-specific worlds. 
On one hand, the training of LM does not need to focus on unrelated domains anymore (e.g., \textit{Books} is one big domain in Amazon but a major focus on Books may not be very helpful for end tasks in \textit{Laptop}); on the other hand, 
although an in-domain corpus may be limited, other relevant domains can share a great amount of knowledge (e.g., Desktop in Table \ref{tbl:example}) to make in-domain corpus more diverse and general.

This paper proposes an extremely simple extension of BERT \cite{devlin2018bert} called DomBERT to learn domain-oriented language models. 
DomBERT divides a mixed-domain corpus by domain tags and learns to re-balance the training examples for the target domain.
Similar to other LMs, we categorize DomBERT as a self-supervised learning model because domain tags naturally exist online and do not require human annotations for a specific task\footnote{Supervised learning needs extra human annotations.}, ranging from Wikipedia, news articles, blog posts, QAs, to customer reviews.
DomBERT simultaneously learns masked language modeling and discovers relevant domains to draw training examples, where the later are computed from domain embeddings learned from an auxiliary task of domain classification.
We apply DomBERT to end tasks in aspect-based sentiment analysis (ABSA) in low-resource settings, demonstrating promising results.

The main contributions of this paper are in 3-fold:
\begin{enumerate}
    \item We propose the task of domain-oriented learning, which aims to learn language models focusing on a target and its relevant domains.
    \item We propose DomBERT, which is an extension of BERT with the capability to draw examples from relevant domains from a pool of diverse domains.
    \item Experimental results demonstrate that DomBERT is promising in low-resource settings for aspect-based sentiment analysis.
\end{enumerate}

\section{Related Work}
Pre-trained language models gain significant improvements over a wide spectrum of NLP tasks, including
ELMo\cite{peters2018deep}, GPT/GPT2\cite{radford2018improving,radford2019language}, BERT\cite{devlin2018bert}, XLNet\cite{yang2019xlnet}, RoBERTa\cite{liu2019roberta}, ALBERT\cite{lan2019albert}, ELECTRA\cite{clark2019electra}.
This paper extends BERT's masked language model (MLM) with domain knowledge learning.
Following RoBERTa, the proposed DomBERT leverages dynamic masking, removes the next sentence prediction (NSP) task (which is proved to have negative effects on pre-trained parameters), and allows for max-length MLM to fully utilize the computational power.
This paper also borrows ALBERT's removal of dropout since pre-trained LM, in general, is an underfitting task that requires more parameters instead of avoiding overfitting.

The proposed domain-oriented learning task can be viewed as one type of transfer learning\cite{pan2009survey}, which learns a transfer strategy implicitly that transfer training examples from relevant (source) domains to the target domain. 
This transfer process is conducted throughout the training process of DomBERT.

The experiment of this paper focuses on aspect-based sentiment analysis (ABSA), which typically requires a lot of domain-specific knowledge.
Reviews serve as a rich resource for sentiment analysis \cite{pang2002thumbs,hu2004mining,liu2012sentiment,liu2015sentiment}.
ABSA aims to turn unstructured reviews into structured fine-grained aspects (such as the ``battery'' or aspect category of a laptop) and their associated opinions (e.g., ``good battery'' is \emph{positive} about the aspect battery).
This paper focuses on three (3) popular tasks in ABSA: aspect extraction (AE), aspect sentiment classification (ASC) \cite{hu2004mining} and end-to-end ABSA (E2E-ABSA) \cite{li2019unified,li2019exploiting}. AE aims to extract aspects (e.g., ``battery''), ASC identifies the polarity for a given aspect (e.g., \emph{positive} for \emph{battery}) and E2E-ABSA is a combination of AE and ASC that detects the aspects and their associated polarities simultaneously.

AE and ASC are two important tasks in sentiment analysis~\cite{pang2002thumbs,liu2015sentiment}. It is different from document or sentence-level sentiment classification (SSC) \cite{pang2002thumbs,kim2014convolutional,he2011self,he2011automatically} as it focuses on fine-grained opinion on each specific aspect \cite{shu-etal-2017-lifelong,xu-etal-2018-double}. It is either studied as a single task or a joint learning end-to-end task together with aspect extraction \cite{wang2017coupled,li2017deep,li2018unified}. 
Most recent works widely use neural networks\cite{dong2014adaptive,nguyen-shirai-2015-phrasernn,li2018transformation}.
For example, memory network \cite{weston2014memory,sukhbaatar2015end} and attention mechanisms are extensively applied to ASC \cite{tang2016aspect,wang2016recursive,wang2016attention,ma2017interactive,chen2017recurrent,ma2017interactive,tay2018learning,he2018effective,liu2018content}. 
ASC is also studied in transfer learning or domain adaptation settings, such as leveraging large-scale corpora that are unlabeled or weakly labeled (e.g., using the overall rating of a review as the label) \cite{xu_bert2019,he-EtAl:2018} and transferring from other tasks/domains \cite{li2018exploiting,wang2018lifelong,wang2018target}. 
Many of these models use handcrafted features, graph structures, lexicons, and complicated neural network architectures to remedy the insufficient training examples from both tasks.
Although these approaches may achieve better performances by manually injecting human knowledge into the model, 
this paper aims to improve ABSA from leveraging unlabeled data via a self-supervised language modeling manner \cite{xu_acl2018,he2018exploiting,xu_bert2019}.

\section{DomBERT}
This section presents DomBERT, which is an extension of BERT for domain knowledge learning.
We adopt post-training of BERT \cite{xu_bert2019} instead of training DomBERT from scratch since post-training on BERT is more efficient\footnote{We aim for single GPU training for all models in this paper and use uncased $\textbf{BERT}_\textbf{BASE}$ given its lower costs of training for academic purpose.}.
Different from \cite{xu_bert2019}, 
the main goal of domain-oriented training is that it leverages both an in-domain corpus and a pool of corpora of source domains.

The goal of DomBERT is to discover relevant domains from the pool of source domains and uses the training examples from relevant source domains for masked language model learning.
As a result, DomBERT has a sampling process over a categorical distribution on all domains (including the target domain) to retrieve relevant domains' examples.
Learning such a distribution needs to detect the domain similarities between all source domains and the target domain.
DomBERT learns an embedding for each domain and computes such similarities.
The domain embeddings are learned from an auxiliary task called domain classification.

\subsection{Domain Classification}
Given a pool of source and target domains, one can easily form a classification task on domain tags.
As such, each text document has its domain label $l$. 
Following RoBERTa\cite{liu2019roberta}'s max-length training examples, 
we pack different texts from the same domain up to the maximum length into a single training example.

We let the number of source domains be $|S|=n$. Then the total number of domains (including the target domain) is $n+1$.
Let $h_{\texttt{[CLS]}}$ denote the hidden state of the \texttt{[CLS]} token of BERT, which indicates the document-level representations of one example.
We first pass this hidden states into a dense layer to reduce the size of hidden states. 
Then we pass this reduced hidden states to a dense layer $\boldsymbol{D} \in \mathbb{R}^{(n+1)*m}$ to compute the logits over all domains $\hat{l}$:
\begin{equation}
\begin{split}
\hat{l} = \boldsymbol{D}\cdot(\boldsymbol{W}\cdot h_{\texttt{[CLS]}} + b),
\end{split}
\end{equation}
where $m$ is the size of the dense layer, $\boldsymbol{D}$, $\boldsymbol{W}$ and $b$ are trainable weights.
Besides a dense layer, $\boldsymbol{D}$ is essentially a concatenation of domain embeddings: $\boldsymbol{D} = d_t \circ d_1 \circ \cdots \circ d_{n} $.
Then we apply cross-entropy loss to the logits and label to obtain the loss of domain classification.
\begin{equation}
\begin{split}
\mathcal{L}_\text{CLS} = \text{CrossEntropyLoss}(\hat{l}, l) .
\end{split}
\end{equation}
To encourage the diversity of domain embeddings, we further compute a regularizer among domain embeddings as following:
\begin{equation}
\begin{split}
\Delta = \frac{1}{|D|^2}||\cos(D, D^T)-I||^2_2 .
\end{split}
\end{equation}
Minimizing this regularizer encourages the learned embeddings to be more orthogonal (thus diverse) to each other.
Finally, we add the loss of domain classification, BERT's masked language model and regularizer together:
\begin{equation}
\begin{split}
\mathcal{L} = \lambda \mathcal{L}_\text{MLM} + (1-\lambda)\mathcal{L}_\text{CLS} + \Delta,
\end{split}
\end{equation}
where $\lambda$ controls the ratio of losses between masked language model and domain classification.

\subsection{Domain Sampler}
As a side product of domain classification, DomBERT has a built-in data sampling process to draw examples from both the target domain and relevant domains for future learning.
This process follows a unified categorical distribution over all domains, which ensures a good amount of examples from both the target domains and relevant domains are sampled.
As such, it is important to always have the target domain $t$ with the highest probability for sampling.

To this end, we use cosine similarity as the similarity function, which has the property to always let $\cos(d_t, d_t)=1$.
For an arbitrary domain $i$, the probability $P_i$ of domain $i$ being sampled is computed from a softmax function over domain similarities as following:
\begin{equation}
P_i = \frac{\exp{(\cos(d_t, d_i) / \tau)}}{\sum^{n+1}_{j = 0} \exp{(\cos(d_t, d_j) / \tau)} },
\end{equation}
where $\tau$ is the temperature \cite{hinton2015distilling} to control the importance of highly-ranked domains vs long-tailed domains. 

To form a mini-batch for the next training step, we sample domains following the categorical distribution of $s \sim P$ up to the batch size and retrieve the next available example from each sampled domain.
As such, we maintain a randomly shuffled queue of examples for each domain. When the examples of one domain are exhausted, a new randomly shuffled queue will be generated for that domain.
As a result, we implement a data sampler that takes $\boldsymbol{D}$ as inputs.

\subsection{Implementation Details}
We adopt the popular transformers framework from hugging face\footnote{\url{https://huggingface.co/transformers/}}, with the following minor improvements.

\noindent \textbf{Early Apply of Labels}:
We refactor the forward computation of BERT MLM by a method called to \underline{e}arly \underline{a}pply of \underline{l}abels (EAL), which leverages labels of MLM in an early stage of forwarding computation to avoid computation for invalid positions. 
Although MLM just uses 15\% of tokens for prediction, the implementations of BERT MLM still computes the logits over the vocabulary for all positions, 
which is a big waste of both GPU computation and memory footprint (because it is expensive to multiply hidden states with word embeddings of vocabulary size).
EAL only uses positions that need prediction when computing logits for each token\footnote{We use \texttt{torch.masked\_select} in PyTorch.}.
This improves the speed of training to 3.2 per second from 2.2 per second for $\textbf{BERT}_\textbf{BASE}$.
A similar method can be applied to compute the cross-entropy based loss for one token (because only the logit for the ground-truth token contributes to the loss), which is potentially useful for almost all tasks with large vocabulary size.

\noindent \textbf{Dropout Removal}:
Following ALBERT\cite{lan2019albert}, we turn off dropout for post-training because BERT is unlikely to overfit to large training corpus.
This gives both a larger capacity of parameters and faster training speed.
The dropout is turned back on during end-task fine-tuning because BERT is typically over-parameterized for end-tasks.

\section{Experiments}
\label{sec:exp}
\subsection{End Task Datasets}
We apply DomBERT to end tasks in aspect-based sentiment analysis from the SemEval dataset, which focusing on \textit{Laptop}, \textit{Restaurant}.
Statistics of datasets for AE, ASC and E2E-ABSA are given in Table \ref{tbl:ae}, \ref{tbl:asc} and \ref{tbl:e2e}, respectively.
For AE, we choose SemEval 2014 Task 4 for laptop and SemEval-2016 Task 5 for restaurant to be consistent with \cite{xu_acl2018} and other previous works.
For ASC, we use SemEval 2014 Task 4 for both laptop and restaurant as existing research frequently uses this version. We use 150 examples from the training set of all these datasets for validation.
For E2E-ABSA, we adopt the formulation of \cite{li2019unified} where the laptop data is from SemEval-2014 task 4 and the restaurant domain is a combination of SemEval 2014-2016. 

\begin{table}[t]
\centering
\scalebox{0.8}{
    \begin{tabular}{l|c|c}
    \hline
    {\bf } &{\bf Laptop} & {\bf Restaurant} \\
    \hline
    Training & \\
    Sentence & 3045 & 2000 \\
    Aspect & 2358 & 1743\\
    \hline\hline
    Testing & \\
    Sentence & 800 & 676 \\
    Aspect & 654 & 622\\
    \hline
    \end{tabular}
}
\caption{Summary of datasets on aspect extraction.}
\label{tbl:ae}
\end{table}

\begin{table}[t]
\centering
\scalebox{0.8}{
    \begin{tabular}{l|c|c}
    \hline
    {\bf } & {\bf Laptop} & {\bf Restaurant}\\
    \hline
    Training & & \\
    Positive & 987 & 2164 \\
    Negative & 866 & 805\\
    Neutral & 460 & 633 \\
    \hline\hline
    Testing & & \\
    Positive & 341 & 728 \\
    Negative & 128 & 196 \\
    Neutral & 169 & 196 \\
    \hline
    \end{tabular}
}
\caption{Summary of datasets on aspect sentiment classification.}
\label{tbl:asc}
\vspace{-5mm}
\end{table}

\begin{table}[t]
\centering
\scalebox{0.8}{
    \begin{tabular}{l|c|c}
    \hline
    {\bf } & {\bf Laptop} & {\bf Restaurant}\\
    \hline
    Training & & \\
    Positive & 987 & 2407 \\
    Negative & 860 & 1035\\
    Neutral & 450 & 664 \\
    \hline\hline
    Testing & & \\
    Positive & 339 & 1524 \\
    Negative & 130 & 500 \\
    Neutral & 165 & 263 \\
    \hline
    \end{tabular}
}
\caption{Summary of datasets on end-to-end aspect-based sentiment analysis.}
\label{tbl:absa}
\vspace{-5mm}
\end{table}

\subsection{Domain Corpus}
\begin{figure}[t]
\centering
\includegraphics[width=1.0\columnwidth]{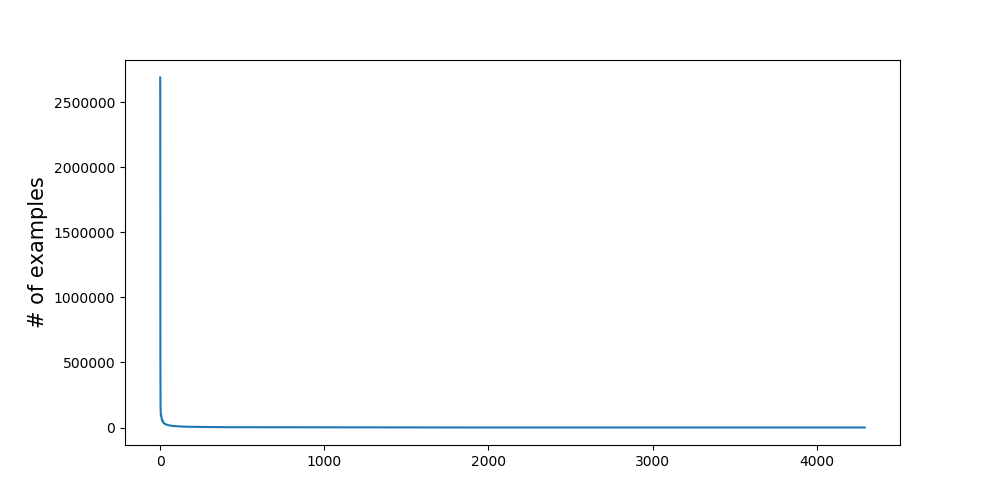}
    \caption{Rank of domains by number of examples.}
\label{fig:domain_cnt}
\vspace{-3mm}
\end{figure}

Based on the domains of end tasks from SemEval dataset, we explore the capability of the large-scale unlabeled corpus from Amazon review datasets\cite{HeMcA16a} and Yelp dataset\footnote{\url{https://www.yelp.com/dataset/challenge}, 2019 version.}.
Following \cite{xu_bert2019}, we select all laptop reviews from the electronics department. This ends with about 100 MB corpus. 
Similarly, we simulate a low-resource setting for restaurants and randomly select about 100 MB reviews tagged as \textit{Restaurants} as their first category from Yelp reviews.
For source domains $S$, we choose all reviews from the 5-core version of Amazon review datasets and all Yelp reviews excluding \textit{Laptop} and \textit{Restaurants}. 
Note that Yelp is not solely about restaurants but has other location-based domains such as car service, bank, theatre etc. 
This ends with a total of $|D|=4680$ domains, and $n=4679$ are source domains. The total size of the corpus is about 20 GB.
The number of examples for each domain is plotted in Figure \ref{fig:domain_cnt}, where the distribution of domains is heavily long-tailed.

\subsection{Hyper-parameters}
\label{sec:hyp}
We adopt $\textbf{BERT}_\textbf{BASE}$ (uncased) as the basis for all experiments due to the limits of computational power in our academic setting and the purpose of making reproducible research.
We choose the hidden size of domain embeddings $m=64$ to ensure the regularizer term in the loss doesn't consume too much GPU memory.
We choose $\tau=0.13$ and $\lambda=0.9$.
We leverage FP16 computation\footnote{\url{https://docs.nvidia.com/deeplearning/sdk/mixed-precision-training/index.html}} to reduce the actual size of tensors on GPU and speed up the training. 
We train with FP16-O2 optimization, which has faster speed and smaller GPU memory footprint compared to O1 optimization.
Due to the uncertainty from the online sampling of DomBERT, we assume the number of training examples per epoch as the number of examples in the target domains.
As a result, we train DomBERT for 400 epochs to get enough training examples from relevant domains.
The full batch size is set to 288 (a multiplication of batch size of 24 and gradient accumulation step 12).
The maximum length of DomBERT is consistent with BERT as 512. 
We use Adamax\cite{kingma2014adam} as the optimizer.
Lastly. the learning rate is to be 5e-5.

\subsection{Compared Methods}
We compare DomBERT with LM-based baselines (that requires no extra supervision from humans such as parsing, fine-grained annotation).

\noindent \textbf{BERT} this is the vanilla $\textbf{BERT}_\textbf{BASE}$ pre-trained model from \cite{devlin2018bert}, which is used to show the performance of BERT without any domain adaption. 

\noindent \textbf{BERT-Review} post-train BERT on all (mixed-domain) Amazon review datasets and Yelp datasets in a similar way of training BERT. Following \cite{liu2019roberta}, we train the whole corpus for 4 epochs, which took about 10 days of training (much longer than DomBERT).

\noindent \textbf{BERT-DK} is a baseline borrowed from \cite{xu_bert2019} that trains an LM per domain. Note that the restaurant domain is trained from 1G of corpus that aligns well with the types of restaurants in SemEval, which is not a low-resource case.
We use this baseline to show that DomBERT can reach competitive performance.

\noindent \textbf{DomBERT} is the model proposed in this paper.

\subsection{Evaluation Metrics}
For AE, we use the standard evaluation scripts come with the SemEval datasets and report the F1 score.
For ASC, we compute both accuracy and Macro-F1 over 3 classes of polarities, where Macro-F1 is the major metric as the imbalanced classes introduce biases on accuracy.~To be consistent with existing research \cite{tang2016aspect}, examples belonging to the \textit{conflict} polarity are dropped due to a very small number of examples.
For E2E-ABSA, we adopt the evaluation script from\cite{li2019unified}, which reports precision, recall, and F1 score on sequence labeling (of combined aspect labels and sentiment polarity).

Results are reported as averages of \textbf{10} runs (10 different random seeds for random batch generation).\footnote{We notice that adopting 5 runs used by existing researches still has a high variance for a fair comparison.} 

\subsection{Result Analysis and Discussion}
Results on different tasks in ABSA exhibit different challenges.

\noindent \textbf{AE}: In Table \ref{tbl:result_ae} We notice that AE is a very domain-specific task. DomBERT further improves the performance of BERT-DK that only uses domain-specific corpus. Note that BERT-DK for restaurant uses 1G of restaurant corpus. But DomBERT's target domain corpus is just 100 MB. So DomBERT further learns domain-specific knowledge from relevant domains. Although Yelp data contain a great portion of restaurant reviews, a mixed-domain training as BERT-Review does not yield enough domain-specific knowledge.

\noindent \textbf{ASC}: ASC is a more domain agnostic task because most of sentiment words are sharable across all domains (e.g., ``good'' and ``bad''). As such, in Table \ref{tbl:result_asc}, we notice ASC for restaurant is more domain-specific than laptop.
DomBERT is worse than BERT-Review in laptop because a 20+ G can learn general-purpose sentiment better. 
BERT-DK is better than DomBERT because a much larger in-domain corpus is more important for performance.

\noindent \textbf{E2E ABSA}: By combining AE and ASC together, E2E ABSA exhibit more domain-specifity, as shown in Table \ref{tbl:result_e2e}. In this case, we can see the full performance of DomBERT because it can learn both general and domain-specific knowledge well. BERT-Review is poor probably because it focuses on irrelevant domains such as \textit{Books}.

We further examine the sampling process of DomBERT. In Table \ref{tbl:top20}, we report top-20 source domains reported by the data sampler at the end of training. The results are closer to our intuition because most domains are very close to laptop and restaurant, respectively.

\begin{table}
    \centering
    \scalebox{0.85}{
        \begin{tabular}{l||c|c}
        \hline
        {\bf Domain} & {\bf Laptop} & {\bf Restaurant} \\
        \hline
        {\bf Methods} & {\bf F1 } & {\bf F1 } \\
        \hline
        BERT\cite{devlin2018bert}  & 79.28 & 74.1 \\
        BERT-Review & 83.64 & 76.20 \\
        BERT-DK\cite{xu_bert2019} & 83.55 & 77.02 \\
        DomBERT & 83.89 & 77.21 \\
        \hline
        \end{tabular}
    }
    \caption{AE in F1.}
\label{tbl:result_ae}
\end{table}

\begin{table}[t]
    \centering
    \scalebox{0.75}{
        \begin{tabular}{l||c c|c c}
        \hline
        {\bf Domain} & {\bf Laptop} & & {\bf Rest.} & \\
        \hline
        {\bf Methods} & \bf{Acc.} & \bf{MF1} & \bf{Acc.} & \bf{MF1} \\
        \hline
        \hline
        BERT\cite{devlin2018bert} & 75.29 & 71.91 & 81.54 & 71.94 \\
        BERT-Review & 78.62 & 75.5 & 83.35 & 74.9 \\
        BERT-DK\cite{xu_bert2019} & 77.01 & 73.72 & 83.96 & 75.45 \\
        DomBERT & 76.72 & 73.46 & 83.14 & 75.00 \\
        \hline
        \end{tabular}
    }
    \caption{ASC in Accuracy and Macro-F1(MF1).}
\label{tbl:result_asc}
\end{table}

\begin{table}[t]
  \centering
  \resizebox{\columnwidth}{!}{
  \begin{tabular}{l|c|c|c|c|c|c}
\hline
 & \multicolumn{3}{|c|}{\bf Laptop} & \multicolumn{3}{|c}{\bf Restaurant}\\
\hline
&P  &R  &F1 &P  &R &F1\\
\hline
Existing Models \\
\hline
\cite{li2019unified} &61.27  &54.89  &57.90  &68.64  &71.01  &69.80\\
\hline
\cite{luo-etal-2019-doer}     &-      &-      &60.35  &-      &-      &72.78\\
\hline
\cite{he_acl2019}      &-      &-      &58.37  &-      &-      &- \\
\hline
\hline
LSTM-CRF \\
\hline
\cite{lample-etal-2016-neural}   &58.61  &50.47  &54.24  &66.10  &66.30  &66.20\\
\hline
\cite{ma-hovy-2016-end}     &58.66  &51.26  &54.71  &61.56  &67.26  &64.29\\
\hline
\cite{2017arXiv170904109L}      &53.31  &59.40  &56.19  &68.46  &64.43  &66.38\\
\hline
\hline
BERT+Linear\cite{li2019exploiting} &62.16  &58.90  &60.43  &71.42  &75.25  &73.22\\
\hline
\hline\hline
BERT\cite{devlin2018bert} & 61.97 & 58.52 & 60.11 & 68.86 & 73.00 & 70.78 \\
BERT-Review & 65.80 & 63.12 & 64.37 & 69.92 & 75.36 & 72.47 \\
BERT-DK\cite{xu_bert2019} & 63.95 & 61.18 & 62.45 & 71.88 & 74.07 & 72.88 \\
DomBERT & 66.96 & 65.58 & 66.21 & 72.17 & 74.96 & 73.45 \\
\hline
\end{tabular}
}
\caption{Results of E2E ABSA: baselines are borrowed from \cite{li2019exploiting}.}
\label{tbl:result_e2e}
\end{table}

\begin{table}[t]
    \centering
    \scalebox{0.65}{
        \begin{tabular}{l|l}
        \hline
        {\bf Laptop} & {\bf Restaurant} \\
        \hline

Tablets & Food \\
Boot Shop (Men) & Coffee \& Tea \\
Laptop \& Netbook Computer Accessories & Bakeries \\
Computers \& Accessories & Bars \\
Microsoft Windows & Nightlife \\
Electronics Warranties & Arts \& Entertainment \\
Desktops & Grocery \\
Antivirus \& Security & Venues \& Event Spaces \\
Aviation Electronics & Lounges \\
Watch Repair & Beer \\
Orthopedists & Casinos \\
Compact Stereos & Hotels \\
Unlocked Cell Phones & Dance Clubs \\
Power Strips & Tea Rooms \\
Mobile Broadband & Pubs \\
Cleaners & Cinema \\
No-Contract Cell Phones & Event Planning \& Services \\
Video Games/PC/Accessories & Sports Bars \\
Antivirus & Specialty Food \\
MP3 Players \& Accessories & Desserts\\
        
        \hline
        \end{tabular}
    }
\caption{Top-20 relevant domains}
\label{tbl:top20}

\end{table}



\section{Conclusions}
This paper investigates the task of domain-oriented learning for language modeling.
It aims to leverage the benefits of both large-scale mixed-domain training and in-domain specific knowledge learning.
We propose a simple extension of BERT called DomBERT, which automatically exploits the power of training corpus from relevant domains for a target domain.
Experimental results demonstrate that the DomBERT is promising in a wide assortment of tasks in aspect-based sentiment analysis.

\bibliographystyle{acl_natbib}
\bibliography{acl2020}

\begin{thebibliography}{52}
\expandafter\ifx\csname natexlab\endcsname\relax\def\natexlab#1{#1}\fi

\bibitem[{Chen et~al.(2017)Chen, Sun, Bing, and Yang}]{chen2017recurrent}
Peng Chen, Zhongqian Sun, Lidong Bing, and Wei Yang. 2017.
\newblock Recurrent attention network on memory for aspect sentiment analysis.
\newblock In \emph{Proceedings of the 2017 conference on empirical methods in
  natural language processing}, pages 452--461.

\bibitem[{Clark et~al.(2019)Clark, Luong, Le, and Manning}]{clark2019electra}
Kevin Clark, Minh-Thang Luong, Quoc~V Le, and Christopher~D Manning. 2019.
\newblock Electra: Pre-training text encoders as discriminators rather than
  generators.
\newblock In \emph{International Conference on Learning Representations}.

\bibitem[{Devlin et~al.(2019)Devlin, Chang, Lee, and
  Toutanova}]{devlin2018bert}
Jacob Devlin, Ming-Wei Chang, Kenton Lee, and Kristina Toutanova. 2019.
\newblock \href {https://doi.org/10.18653/v1/N19-1423} {{BERT}: Pre-training of
  deep bidirectional transformers for language understanding}.
\newblock In \emph{Proceedings of the 2019 Conference of the North {A}merican
  Chapter of the Association for Computational Linguistics: Human Language
  Technologies, Volume 1 (Long and Short Papers)}, pages 4171--4186,
  Minneapolis, Minnesota. Association for Computational Linguistics.

\bibitem[{Dong et~al.(2014)Dong, Wei, Tan, Tang, Zhou, and
  Xu}]{dong2014adaptive}
Li~Dong, Furu Wei, Chuanqi Tan, Duyu Tang, Ming Zhou, and Ke~Xu. 2014.
\newblock Adaptive recursive neural network for target-dependent twitter
  sentiment classification.
\newblock In \emph{Proceedings of the 52nd annual meeting of the association
  for computational linguistics (volume 2: Short papers)}, volume~2, pages
  49--54.

\bibitem[{Gururangan et~al.(2020)Gururangan, Marasović, Swayamdipta, Lo,
  Beltagy, Downey, and Smith}]{dontstoppretraining2020}
Suchin Gururangan, Ana Marasović, Swabha Swayamdipta, Kyle Lo, Iz~Beltagy,
  Doug Downey, and Noah~A. Smith. 2020.
\newblock Don't stop pretraining: Adapt language models to domains and tasks.
\newblock In \emph{Proceedings of ACL}.

\bibitem[{He et~al.({\natexlab{a}})He, Lee, Ng, and Dahlmeier}]{he-EtAl:2018}
Ruidan He, Wee~Sun Lee, Hwee~Tou Ng, and Daniel Dahlmeier. {\natexlab{a}}.
\newblock Exploiting document knowledge for aspect-level sentiment
  classification.
\newblock In \emph{Proceedings of the 56th Annual Meeting of the Association
  for Computational Linguistics}. Association for Computational Linguistics.

\bibitem[{He et~al.({\natexlab{b}})He, Lee, Ng, and Dahlmeier}]{he_acl2019}
Ruidan He, Wee~Sun Lee, Hwee~Tou Ng, and Daniel Dahlmeier. {\natexlab{b}}.
\newblock An interactive multi-task learning network for end-to-end
  aspect-based sentiment analysis.
\newblock In \emph{Proceedings of the 57th Annual Meeting of the Association
  for Computational Linguistics}. Association for Computational Linguistics.

\bibitem[{He et~al.(2018{\natexlab{a}})He, Lee, Ng, and
  Dahlmeier}]{he2018effective}
Ruidan He, Wee~Sun Lee, Hwee~Tou Ng, and Daniel Dahlmeier. 2018{\natexlab{a}}.
\newblock Effective attention modeling for aspect-level sentiment
  classification.
\newblock In \emph{Proceedings of the 27th International Conference on
  Computational Linguistics}, pages 1121--1131.

\bibitem[{He et~al.(2018{\natexlab{b}})He, Lee, Ng, and
  Dahlmeier}]{he2018exploiting}
Ruidan He, Wee~Sun Lee, Hwee~Tou Ng, and Daniel Dahlmeier. 2018{\natexlab{b}}.
\newblock Exploiting document knowledge for aspect-level sentiment
  classification.
\newblock \emph{arXiv preprint arXiv:1806.04346}.

\bibitem[{He and McAuley(2016)}]{HeMcA16a}
Ruining He and Julian McAuley. 2016.
\newblock Ups and downs: Modeling the visual evolution of fashion trends with
  one-class collaborative filtering.
\newblock In \emph{World Wide Web}.

\bibitem[{He et~al.(2011)He, Lin, and Alani}]{he2011automatically}
Yulan He, Chenghua Lin, and Harith Alani. 2011.
\newblock Automatically extracting polarity-bearing topics for cross-domain
  sentiment classification.
\newblock In \emph{Proceedings of the 49th Annual Meeting of the Association
  for Computational Linguistics: Human Language Technologies-Volume 1}, pages
  123--131. Association for Computational Linguistics.

\bibitem[{He and Zhou(2011)}]{he2011self}
Yulan He and Deyu Zhou. 2011.
\newblock Self-training from labeled features for sentiment analysis.
\newblock \emph{Information Processing \& Management}, 47(4):606--616.

\bibitem[{Hinton et~al.(2015)Hinton, Vinyals, and Dean}]{hinton2015distilling}
Geoffrey Hinton, Oriol Vinyals, and Jeff Dean. 2015.
\newblock Distilling the knowledge in a neural network.
\newblock \emph{arXiv preprint arXiv:1503.02531}.

\bibitem[{Hu and Liu(2004)}]{hu2004mining}
Minqing Hu and Bing Liu. 2004.
\newblock Mining and summarizing customer reviews.
\newblock In \emph{Proceedings of the tenth ACM SIGKDD international conference
  on Knowledge discovery and data mining}, pages 168--177. ACM.

\bibitem[{Kim(2014)}]{kim2014convolutional}
Yoon Kim. 2014.
\newblock Convolutional neural networks for sentence classification.
\newblock In \emph{Proceedings of the 2014 Conference on Empirical Methods in
  Natural Language Processing (EMNLP)}, pages 1746--1751.

\bibitem[{Kingma and Ba(2014)}]{kingma2014adam}
Diederik~P Kingma and Jimmy Ba. 2014.
\newblock Adam: A method for stochastic optimization.
\newblock \emph{arXiv preprint arXiv:1412.6980}.

\bibitem[{Lample et~al.(2016)Lample, Ballesteros, Subramanian, Kawakami, and
  Dyer}]{lample-etal-2016-neural}
Guillaume Lample, Miguel Ballesteros, Sandeep Subramanian, Kazuya Kawakami, and
  Chris Dyer. 2016.
\newblock \href {https://doi.org/10.18653/v1/N16-1030} {Neural architectures
  for named entity recognition}.
\newblock In \emph{Proceedings of the 2016 Conference of the North {A}merican
  Chapter of the Association for Computational Linguistics: Human Language
  Technologies}, pages 260--270, San Diego, California. Association for
  Computational Linguistics.

\bibitem[{Lan et~al.(2019)Lan, Chen, Goodman, Gimpel, Sharma, and
  Soricut}]{lan2019albert}
Zhenzhong Lan, Mingda Chen, Sebastian Goodman, Kevin Gimpel, Piyush Sharma, and
  Radu Soricut. 2019.
\newblock Albert: A lite bert for self-supervised learning of language
  representations.
\newblock In \emph{International Conference on Learning Representations}.

\bibitem[{Li et~al.(2018{\natexlab{a}})Li, Bing, Lam, and
  Shi}]{li2018transformation}
Xin Li, Lidong Bing, Wai Lam, and Bei Shi. 2018{\natexlab{a}}.
\newblock Transformation networks for target-oriented sentiment classification.
\newblock \emph{arXiv preprint arXiv:1805.01086}.

\bibitem[{Li et~al.(2019{\natexlab{a}})Li, Bing, Li, and Lam}]{li2019unified}
Xin Li, Lidong Bing, Piji Li, and Wai Lam. 2019{\natexlab{a}}.
\newblock A unified model for opinion target extraction and target sentiment
  prediction.
\newblock In \emph{Proceedings of the AAAI Conference on Artificial
  Intelligence}, volume~33, pages 6714--6721.

\bibitem[{Li et~al.(2019{\natexlab{b}})Li, Bing, Li, and Lam}]{li2018unified}
Xin Li, Lidong Bing, Piji Li, and Wai Lam. 2019{\natexlab{b}}.
\newblock A unified model for opinion target extraction and target sentiment
  prediction.
\newblock In \emph{Proceedings of the AAAI Conference on Artificial
  Intelligence}, volume~33, pages 6714--6721.

\bibitem[{Li et~al.(2019{\natexlab{c}})Li, Bing, Zhang, and
  Lam}]{li2019exploiting}
Xin Li, Lidong Bing, Wenxuan Zhang, and Wai Lam. 2019{\natexlab{c}}.
\newblock Exploiting bert for end-to-end aspect-based sentiment analysis.
\newblock \emph{arXiv preprint arXiv:1910.00883}.

\bibitem[{Li and Lam(2017)}]{li2017deep}
Xin Li and Wai Lam. 2017.
\newblock Deep multi-task learning for aspect term extraction with memory
  interaction.
\newblock In \emph{Proceedings of the 2017 Conference on Empirical Methods in
  Natural Language Processing}, pages 2886--2892.

\bibitem[{Li et~al.(2018{\natexlab{b}})Li, Wei, Zhang, Zhang, Li, and
  Yang}]{li2018exploiting}
Zheng Li, Ying Wei, Yu~Zhang, Xiang Zhang, Xin Li, and Qiang Yang.
  2018{\natexlab{b}}.
\newblock Exploiting coarse-to-fine task transfer for aspect-level sentiment
  classification.
\newblock \emph{arXiv preprint arXiv:1811.10999}.

\bibitem[{Liu(2012)}]{liu2012sentiment}
Bing Liu. 2012.
\newblock Sentiment analysis and opinion mining.
\newblock \emph{Synthesis lectures on human language technologies},
  5(1):1--167.

\bibitem[{Liu(2015)}]{liu2015sentiment}
Bing Liu. 2015.
\newblock \emph{Sentiment analysis: Mining opinions, sentiments, and emotions}.
\newblock Cambridge University Press.

\bibitem[{{Liu} et~al.(2018){Liu}, {Shang}, {Xu}, {Ren}, {Gui}, {Peng}, and
  {Han}}]{2017arXiv170904109L}
L.~{Liu}, J.~{Shang}, F.~{Xu}, X.~{Ren}, H.~{Gui}, J.~{Peng}, and J.~{Han}.
  2018.
\newblock {Empower Sequence Labeling with Task-Aware Neural Language Model}.
\newblock In \emph{AAAI}.

\bibitem[{Liu et~al.(2018)Liu, Zhang, Zeng, Huang, and Wu}]{liu2018content}
Qiao Liu, Haibin Zhang, Yifu Zeng, Ziqi Huang, and Zufeng Wu. 2018.
\newblock Content attention model for aspect based sentiment analysis.
\newblock In \emph{Proceedings of the 2018 World Wide Web Conference on World
  Wide Web}, pages 1023--1032. International World Wide Web Conferences
  Steering Committee.

\bibitem[{Liu et~al.(2019)Liu, Ott, Goyal, Du, Joshi, Chen, Levy, Lewis,
  Zettlemoyer, and Stoyanov}]{liu2019roberta}
Yinhan Liu, Myle Ott, Naman Goyal, Jingfei Du, Mandar Joshi, Danqi Chen, Omer
  Levy, Mike Lewis, Luke Zettlemoyer, and Veselin Stoyanov. 2019.
\newblock Roberta: A robustly optimized bert pretraining approach.
\newblock \emph{arXiv preprint arXiv:1907.11692}.

\bibitem[{Luo et~al.(2019)Luo, Li, Liu, and Zhang}]{luo-etal-2019-doer}
Huaishao Luo, Tianrui Li, Bing Liu, and Junbo Zhang. 2019.
\newblock \href {https://doi.org/10.18653/v1/P19-1056} {{DOER}: Dual
  cross-shared {RNN} for aspect term-polarity co-extraction}.
\newblock In \emph{Proceedings of the 57th Annual Meeting of the Association
  for Computational Linguistics}, pages 591--601, Florence, Italy. Association
  for Computational Linguistics.

\bibitem[{Ma et~al.(2017)Ma, Li, Zhang, and Wang}]{ma2017interactive}
Dehong Ma, Sujian Li, Xiaodong Zhang, and Houfeng Wang. 2017.
\newblock Interactive attention networks for aspect-level sentiment
  classification.
\newblock \emph{arXiv preprint arXiv:1709.00893}.

\bibitem[{Ma and Hovy(2016)}]{ma-hovy-2016-end}
Xuezhe Ma and Eduard Hovy. 2016.
\newblock \href {https://doi.org/10.18653/v1/P16-1101} {End-to-end sequence
  labeling via bi-directional {LSTM}-{CNN}s-{CRF}}.
\newblock In \emph{Proceedings of the 54th Annual Meeting of the Association
  for Computational Linguistics (Volume 1: Long Papers)}, pages 1064--1074,
  Berlin, Germany. Association for Computational Linguistics.

\bibitem[{Nguyen and Shirai(2015)}]{nguyen-shirai-2015-phrasernn}
Thien~Hai Nguyen and Kiyoaki Shirai. 2015.
\newblock \href {https://doi.org/10.18653/v1/D15-1298} {{P}hrase{RNN}: Phrase
  recursive neural network for aspect-based sentiment analysis}.
\newblock In \emph{Proceedings of the 2015 Conference on Empirical Methods in
  Natural Language Processing}, pages 2509--2514, Lisbon, Portugal. Association
  for Computational Linguistics.

\bibitem[{Pan and Yang(2009)}]{pan2009survey}
Sinno~Jialin Pan and Qiang Yang. 2009.
\newblock A survey on transfer learning.
\newblock \emph{IEEE Transactions on knowledge and data engineering},
  22(10):1345--1359.

\bibitem[{Pang et~al.(2002)Pang, Lee, and Vaithyanathan}]{pang2002thumbs}
Bo~Pang, Lillian Lee, and Shivakumar Vaithyanathan. 2002.
\newblock Thumbs up?: sentiment classification using machine learning
  techniques.
\newblock In \emph{Proceedings of the ACL-02 conference on Empirical methods in
  natural language processing-Volume 10}, pages 79--86. Association for
  Computational Linguistics.

\bibitem[{Peters et~al.(2018)Peters, Neumann, Iyyer, Gardner, Clark, Lee, and
  Zettlemoyer}]{peters2018deep}
Matthew~E Peters, Mark Neumann, Mohit Iyyer, Matt Gardner, Christopher Clark,
  Kenton Lee, and Luke Zettlemoyer. 2018.
\newblock Deep contextualized word representations.
\newblock In \emph{Proceedings of NAACL-HLT}, pages 2227--2237.

\bibitem[{Radford et~al.(2018)Radford, Narasimhan, Salimans, and
  Sutskever}]{radford2018improving}
Alec Radford, Karthik Narasimhan, Tim Salimans, and Ilya Sutskever. 2018.
\newblock Improving language understanding by generative pre-training.
\newblock \emph{URL
  https://s3-us-west-2.amazonaws.com/openai-assets/research-covers/languageunsupervised/language
  understanding paper.pdf}.

\bibitem[{Radford et~al.(2019)Radford, Wu, Child, Luan, Amodei, and
  Sutskever}]{radford2019language}
Alec Radford, Jeffrey Wu, Rewon Child, David Luan, Dario Amodei, and Ilya
  Sutskever. 2019.
\newblock Language models are unsupervised multitask learners.
\newblock \emph{OpenAI Blog}, 1(8):9.

\bibitem[{Shu et~al.(2017)Shu, Xu, and Liu}]{shu-etal-2017-lifelong}
Lei Shu, Hu~Xu, and Bing Liu. 2017.
\newblock \href {https://doi.org/10.18653/v1/P17-2023} {Lifelong learning {CRF}
  for supervised aspect extraction}.
\newblock In \emph{Proceedings of the 55th Annual Meeting of the Association
  for Computational Linguistics (Volume 2: Short Papers)}, pages 148--154,
  Vancouver, Canada. Association for Computational Linguistics.

\bibitem[{Sukhbaatar et~al.(2015)Sukhbaatar, Weston, Fergus
  et~al.}]{sukhbaatar2015end}
Sainbayar Sukhbaatar, Jason Weston, Rob Fergus, et~al. 2015.
\newblock End-to-end memory networks.
\newblock In \emph{Advances in neural information processing systems}, pages
  2440--2448.

\bibitem[{Tang et~al.(2016)Tang, Qin, and Liu}]{tang2016aspect}
Duyu Tang, Bing Qin, and Ting Liu. 2016.
\newblock Aspect level sentiment classification with deep memory network.
\newblock \emph{arXiv preprint arXiv:1605.08900}.

\bibitem[{Tay et~al.(2018)Tay, Tuan, and Hui}]{tay2018learning}
Yi~Tay, Luu~Anh Tuan, and Siu~Cheung Hui. 2018.
\newblock Learning to attend via word-aspect associative fusion for
  aspect-based sentiment analysis.
\newblock In \emph{Thirty-Second AAAI Conference on Artificial Intelligence}.

\bibitem[{Wang et~al.(2018{\natexlab{a}})Wang, Lv, Mazumder, Fei, and
  Liu}]{wang2018lifelong}
Shuai Wang, Guangyi Lv, Sahisnu Mazumder, Geli Fei, and Bing Liu.
  2018{\natexlab{a}}.
\newblock Lifelong learning memory networks for aspect sentiment
  classification.
\newblock In \emph{2018 IEEE International Conference on Big Data (Big Data)},
  pages 861--870. IEEE.

\bibitem[{Wang et~al.(2018{\natexlab{b}})Wang, Mazumder, Liu, Zhou, and
  Chang}]{wang2018target}
Shuai Wang, Sahisnu Mazumder, Bing Liu, Mianwei Zhou, and Yi~Chang.
  2018{\natexlab{b}}.
\newblock Target-sensitive memory networks for aspect sentiment classification.
\newblock In \emph{Proceedings of the 56th Annual Meeting of the Association
  for Computational Linguistics (Volume 1: Long Papers)}, pages 957--967.

\bibitem[{Wang et~al.(2016{\natexlab{a}})Wang, Pan, Dahlmeier, and
  Xiao}]{wang2016recursive}
Wenya Wang, Sinno~Jialin Pan, Daniel Dahlmeier, and Xiaokui Xiao.
  2016{\natexlab{a}}.
\newblock Recursive neural conditional random fields for aspect-based sentiment
  analysis.
\newblock \emph{arXiv preprint arXiv:1603.06679}.

\bibitem[{Wang et~al.(2017)Wang, Pan, Dahlmeier, and Xiao}]{wang2017coupled}
Wenya Wang, Sinno~Jialin Pan, Daniel Dahlmeier, and Xiaokui Xiao. 2017.
\newblock Coupled multi-layer attentions for co-extraction of aspect and
  opinion terms.
\newblock In \emph{Thirty-First AAAI Conference on Artificial Intelligence}.

\bibitem[{Wang et~al.(2016{\natexlab{b}})Wang, Huang, Zhao
  et~al.}]{wang2016attention}
Yequan Wang, Minlie Huang, Li~Zhao, et~al. 2016{\natexlab{b}}.
\newblock Attention-based lstm for aspect-level sentiment classification.
\newblock In \emph{Proceedings of the 2016 conference on empirical methods in
  natural language processing}, pages 606--615.

\bibitem[{Weston et~al.(2014)Weston, Chopra, and Bordes}]{weston2014memory}
Jason Weston, Sumit Chopra, and Antoine Bordes. 2014.
\newblock Memory networks.
\newblock \emph{arXiv preprint arXiv:1410.3916}.

\bibitem[{Xu et~al.(2018{\natexlab{a}})Xu, Liu, Shu, and Yu}]{xu_acl2018}
Hu~Xu, Bing Liu, Lei Shu, and Philip~S. Yu. 2018{\natexlab{a}}.
\newblock Double embeddings and cnn-based sequence labeling for aspect
  extraction.
\newblock In \emph{Proceedings of the 56th Annual Meeting of the Association
  for Computational Linguistics}. Association for Computational Linguistics.

\bibitem[{Xu et~al.(2018{\natexlab{b}})Xu, Liu, Shu, and
  Yu}]{xu-etal-2018-double}
Hu~Xu, Bing Liu, Lei Shu, and Philip~S. Yu. 2018{\natexlab{b}}.
\newblock \href {https://doi.org/10.18653/v1/P18-2094} {Double embeddings and
  {CNN}-based sequence labeling for aspect extraction}.
\newblock In \emph{Proceedings of the 56th Annual Meeting of the Association
  for Computational Linguistics (Volume 2: Short Papers)}, pages 592--598,
  Melbourne, Australia. Association for Computational Linguistics.

\bibitem[{Xu et~al.(2019)Xu, Liu, Shu, and Yu}]{xu_bert2019}
Hu~Xu, Bing Liu, Lei Shu, and Philip~S. Yu. 2019.
\newblock Bert post-training for review reading comprehension and aspect-based
  sentiment analysis.
\newblock In \emph{Proceedings of the 2019 Conference of the North American
  Chapter of the Association for Computational Linguistics}.

\bibitem[{Yang et~al.(2019)Yang, Dai, Yang, Carbonell, Salakhutdinov, and
  Le}]{yang2019xlnet}
Zhilin Yang, Zihang Dai, Yiming Yang, Jaime Carbonell, Russ~R Salakhutdinov,
  and Quoc~V Le. 2019.
\newblock Xlnet: Generalized autoregressive pretraining for language
  understanding.
\newblock In \emph{Advances in neural information processing systems}, pages
  5754--5764.

\end{thebibliography}

\end{document}